\documentclass[letterpaper, 10 pt, conference]{ieeeconf}  

\usepackage{cite}
\usepackage{amsmath,amssymb,amsfonts}
\usepackage{algorithm}
\usepackage[noend]{algpseudocode}
\usepackage{subcaption}
\usepackage{graphicx}
\usepackage{physics}
\usepackage{textcomp}
\usepackage{hyperref}
\usepackage{gensymb}
\usepackage{multirow}
\let\labelindent\relax
\usepackage{enumitem}
\usepackage[font=small,labelfont=bf]{caption}
\usepackage[table,xcdraw]{xcolor}
\usepackage{adjustbox}
\usepackage[font=scriptsize]{caption} 

\title{\LARGE \bf
COVINS-G: A Generic Back-end for Collaborative Visual-Inertial SLAM
}

\author{Manthan Patel, Marco Karrer, Philipp B\"anninger and Margarita Chli \\
Vision for Robotics Lab, ETH Zurich, Switzerland\\
}

\begin{document}

\maketitle
\thispagestyle{empty}
\pagestyle{empty}

\begin{abstract}
Collaborative SLAM is at the core of perception in multi-robot systems as it enables the co-localization of the team of robots in a common reference frame, which is of vital importance for any coordination amongst them.
The paradigm of a centralized architecture is well established, with the robots (i.e. agents) running Visual-Inertial Odometry (VIO) onboard while communicating relevant data, such as e.g. Keyframes (KFs), to a central back-end (i.e. server), which then merges and optimizes the joint maps of the agents.
While these frameworks have proven to be successful, their capability and performance are highly dependent on the choice of the VIO front-end, thus limiting their flexibility.
In this work, we present COVINS-G, a generalized back-end building upon the COVINS \cite{covins} framework, enabling the compatibility of the server-back-end with any arbitrary VIO front-end, including, for example, off-the-shelf cameras with odometry capabilities, such as the Realsense T265.
The COVINS-G back-end deploys a multi-camera relative pose estimation algorithm for computing the loop-closure constraints allowing the system to work purely on 2D image data.
In the experimental evaluation, we show on-par accuracy with state-of-the-art multi-session and collaborative SLAM systems, while demonstrating the flexibility and generality of our approach by employing different front-ends onboard collaborating agents within the same mission.
The COVINS-G codebase along with a generalized front-end wrapper to allow any existing VIO front-end to be readily used in combination with the proposed collaborative back-end is open-sourced. 

\textit{Video--} \href{https://youtu.be/FoJfXCfaYDw}{https://youtu.be/FoJfXCfaYDw}
\end{abstract}

\vspace{0pt}

\section{Introduction}
\label{sec:introduction}
The promise of automating tedious tasks and keeping humans away from hazardous environments has driven the development of robotic systems to reach tremendous capabilities.
With the increased maturity of single agent systems, the interest in teams of robots collaborating with each other has been steadily rising.
The use of such a team of robots collaborating towards a common goal promises to increase the efficiency and robustness of a mission by distributing the tasks among the participating agents and has been proposed for a variety of different tasks, such as in search-and-rescue missions \cite{srr}, archaeological mapping \cite{col_arch_map}, precision agriculture \cite{agri-robots}, and surveillance \cite{surveillance}.
Sharing information across the agents not only enables the robots to perform a task faster but also enables the individual agents to make better-informed decisions as they can profit from information beyond their own gathered experience, as shown in \cite{bartolomei2020multi}.

\begin{figure}
    \centering
    \includegraphics[width=0.48\textwidth]{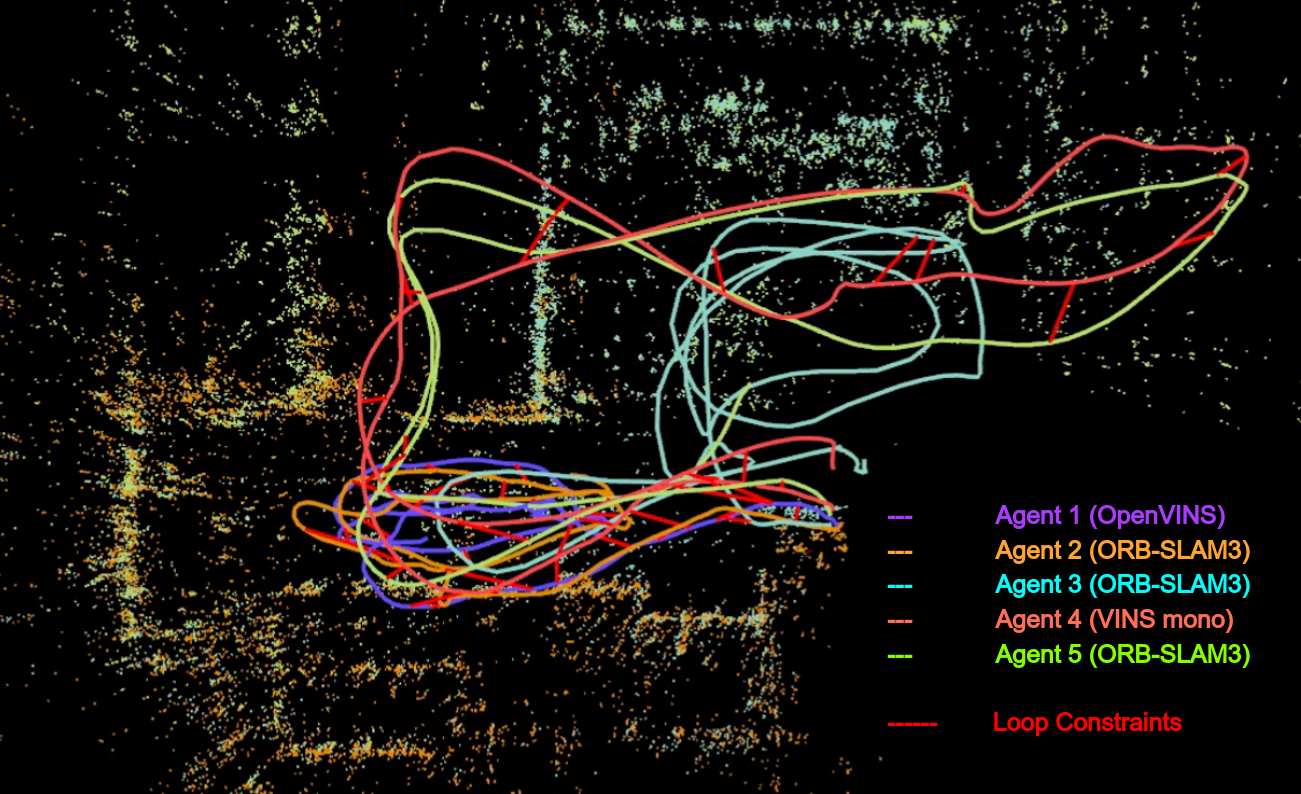}
    \caption{Collaborative SLAM estimate for 5 agents (EuRoC MH Dataset) running different VIO front-ends. The COVINS-G back-end does not require map points (shown here only for visualization) and thus, is compatible with any arbitrary VIO front-end.
    }
    \label{fig:irchel_5ag}    
    \vspace{-20pt}
\end{figure}

However, in order for the robots to work towards any higher-level goal, they need to be aware of their surroundings and their pose in their workspace.
Moreover, for the robots to be able to collaborate, the knowledge of the pose of all other robots within the team is crucial.
The use of external sensors, such as GPS or motion capture systems can provide such a shared reference frame enabling the coordination of the robotic team, however, for many practical applications, such data is not reliable or simply not available in the first place.
For example, the uncertainty of GPS measurements can be in the order of tens of meters close to larger structures (e.g. within a city) or not available at all inside buildings or underground.
In order to remove dependencies on external sensors, research into Simultaneous Localization And Mapping (SLAM) has made significant progress.
In particular, the use of cameras and Inertial Measurement Units (IMUs) for visual-inertial SLAM has proven to provide robustness and accuracy, which led to their deployment onboard products in the market already.
With the increasing maturity of single-agent vision-based SLAM techniques, the extension towards multi-agent SLAM as a core enabler for robotic collaboration in real-world scenarios has been increasingly gaining interest, sparking a variety of works addressing multi-agent SLAM \cite{multi-uav-slam, cvi-slam, ccm-slam, door-slam, kimera-multi, covins, karrer2021}. 
While the capabilities and robustness of the developed systems have been steadily increasing, due to their tailored architectures, their modularity is often sacrificed. 
As a result, any exchange and modification of the front-end onboard such systems most often requires significant effort to adapt the back-end to the structure.
In this spirit, this work addresses this issue by proposing a generic back-end solution built on top of the architecture of \cite{covins}.
As the proposed approach requires only 2D features, it is agnostic to the front-end running onboard each agent and even allows mixing different front-end algorithms on the different agents during the same mission as illustrated in Figure \ref{fig:irchel_5ag}.
In summary, the contributions of this work are the following:
\begin{itemize}[leftmargin=10pt]
    \item a generalized collaborative SLAM back-end, which requires only 2D keypoints and a pose estimate to fuse the estimates from multiple agents, enabling the use of any arbitrary VIO and stereo front-ends onboard each agent, 
    \item a publicly available codebase\footnote{\href{https://github.com/VIS4ROB-lab/covins}{https://github.com/VIS4ROB-lab/covins}}, which is integrated with the framework of \cite{covins}. Furthermore, a front-end wrapper is provided, to support any off-the-shelf front-end, and
    \item an extensive evaluation of the proposed back-end on both the EuRoC dataset \cite{euroc} as well as newly collected datasets. Our evaluation reveals the flexibility of the proposed approach, using and combining different types of front-ends onboard collaborating agents within the same mission.
\end{itemize}

\vspace{0pt}
\section{Related Work}
\label{sec:relatedwork}
The capability to process multiple trajectories sequentially ({\it{aka}} multi-session capability) can be seen as a special case of collaborative SLAM.
Recent SLAM systems, such as ORB-SLAM3\cite{ORBSLAM3_TRO} and VINS-mono \cite{vins-mono} have such multi-session capabilities, which enables them to achieve joint pose and scene estimates similar to collaborative SLAM estimates.
While these approaches achieve greater accuracy and robustness when compared to single-agent SLAM, they are not designed to be used in real-time applications where multiple agents are operating at the same time.

In multi-agent SLAM literature, the classification of the systems is generally made into decentralized and centralized architectures.
One of the first decentralized approaches to multi-agent SLAM is DDF-SAM \cite{ddfsam}, which communicates and propagates condensed local graphs between the robots to distribute the information.
Combining efficient decentralized place recognition \cite{Cieslewski:Scaramuzza:MRS2017} and with a Gauss-Seidel based distributed pose graph optimization \cite{Coudhary:etal:ICRA2016}, in \cite{Cieslewski:etal:ICRA2018} a data-efficient and complete decentralized visual SLAM approach was proposed. 
In \cite{dist-coslam}, a monocular vision-only distributed SLAM for mapping large-scale environments is presented. 
The recent works \cite{door-slam, kimera-multi} both make use of distributed pose graph optimization schemes along with a robust mechanism for identifying and rejecting incorrect loop closures. 
While these distributed SLAM approaches have advantages in terms of scalability, in general, as the information is also distributed the extent of collaboration is limited in order to keep the communication requirements feasible.

On the other hand, in centralized systems, all relevant information passes through a central node.
In \cite{PG-slam}, the authors present a back-end for real-time multi-robot collaborative SLAM where the server combines the local pose graphs obtained from different agents into a global pose graph.
In MOARSLAM \cite{moarslam}, each agent runs a full SLAM pipeline onboard, while the server is used to store the agents' maps and perform map merges across them.
As the agents perform all the computationally expensive tasks, in particular global optimization, this approach is not well-suited for resource-constrained platforms.
On the other end of the spectrum is  C${}^2$TAM \cite{riazuelo2014c}, which offloads all tasks onto a server platform, except pose tracking.
While this allows for very limited computational load onboard the agents, it limits the autonomy of the agents, as a loss of connection to the server eventually causes the onboard tracking to fail.
A middle ground between MOARSLAM and C${}^2$TAM was proposed in \cite{multi-uav-slam}, which introduces a system architecture enabling the agents to function autonomously, but is still able to offload heavy computations to a server and crucially, enable two-way information flow between the agents and the server.
This work was extended in CCM-SLAM \cite{ccm-slam}, shown to perform in real-time on real data, proposing redundancy detection that enabled scalability, which is key in large-scale missions.
Pushing for the incorporation of inertial cues onboard each agent, aside from the monocular cues, CVI-SLAM \cite{cvi-slam} was demonstrated to achieve higher accuracy and metrically scaled SLAM estimates aligned with the gravity direction -- which are core to robotic autonomy in reality.
Most recently, COVINS \cite{covins} was proposed, revisiting the most important components for centralized collaborative SLAM, shown to achieve significant gains in accuracy, while pushing the scalability of the architecture to up to 12 participating agents.

Despite the improved performance, however, one of the major limitations of COVINS is that this performance is highly dependent on the choice of the VIO front-end employed onboard the agents.
For example, using an ORB-SLAM3\cite{ORBSLAM3_TRO} front-end with the COVINS back-end gives an exceptional performance, but the performance drops significantly when using an alternative state-of-the-art VIO front-end, such as VINS-mono \cite{vins-mono}.
This is mainly caused due to the reliance of COVINS on large numbers of highly accurate map points for closing loops, which holds for ORB-SLAM3, for example, but not for VINS-mono. 
By utilizing a generic multi-camera relative pose solver for map fusion and loop-closures we break the dependency on highly accurate map points and instead perform the operations using only 2D image observations.
This does not only enable COVINS-G to perform well with virtually any VIO frontend, but also permits the usage of more powerful image features, which in return, allows to handle drastic view-point changes where previous systems failed.


\section{Methodology}
\label{sec:methodology}
In this section, we first provide an overview of the overall architecture in Section \ref{sec:architecture}.
Since our system is closely inspired by the COVINS framework, we then focus on the individual modules of the whole architecture, which are directly impacted by the contributions of this work.

\subsection{System Overview}
\label{sec:architecture}
\begin{figure}
    \centering
    \includegraphics[width=0.47\textwidth]{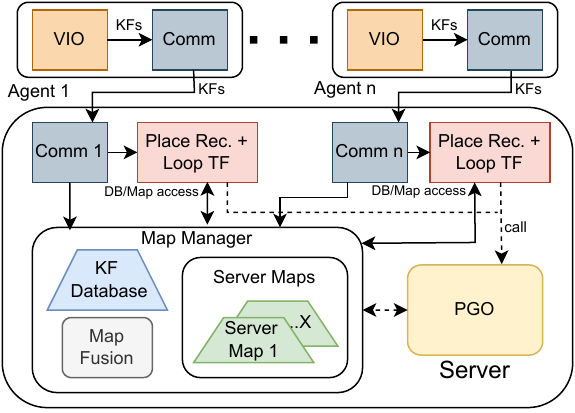}
    \caption{Overview of the COVINS-G system architecture.
    }
    \label{fig:architecture}    
    \vspace{-19pt}
\end{figure}

A summary of our system architecture can be seen in Figure \ref{fig:architecture}.
The system consists of $n$ individual agents, which all run their own local VIO independently and are able to communicate their Keyframe (KF) information to a central server.
The communication module is based on the approach of \cite{covins}, and is shown to be resilient to potential message losses as well as network delays and occasional bottlenecks in the bandwidth.
However, owing to the fact that each agent is running an independent VIO, even a complete loss in connection does not violate the autonomy of an agent.

The central server, on the other hand, is responsible for maintaining the data of the individual agents, fusing the information from the agent, and carrying out computationally expensive tasks, such as global optimization.
After decoding the received KF information, for every KF a visual descriptor based on the Bag-Of-Words (BoW) approach gets computed and added to the KF Database containing the visual descriptor for all the collected KFs.
For every incoming KF, a place-recognition query is performed to find the closest visual matches (candidate KFs).
Based on the visual similarity, it is attempted to compute the transformation between the newly received KF and the candidate KFs.
Upon successful computation of the transformation between the KFs, it is used to either perform a loop closure within a map or if it is across multiple maps, these get fused based on the computed transformation.

At the start of a mission, each of the agents is initialized separately and has a dedicated map on the server.
In our system, the map closely resembles a Pose Graph, where nodes correspond to KFs and edges are based on the relative poses from the odometry and loop closures.
As soon as a loop is found across two agents, their maps get merged together by transforming the map of one agent into the coordinate frame of the other based on the estimated loop transformation.
Upon detection of any loop, a Pose Graph Optimization (PGO) is carried out over all connected Maps.

As opposed to COVINS, which strongly relies on having a well-estimated and consistent set of map points, in our approach we only operate on 2D keypoint information.
This opens up the possibility to use and combine all sorts of front-ends, even if no map points are accessible at all (e.g. in the case of an off-the-shelf tracking sensor such as the T265).
In order to be able to compute the loop constraints without access to map points, we make use of a multi-camera relative pose estimation algorithm \cite{17PT} by treating neighboring KFs as a multi-camera system.

\subsection{Loop Closure and Map Fusion}
\label{sec:loop_closure}
The process of closing loops consists of two main processes; first, suitable candidates for the current query KF are detected, and second, the candidates get geometrically verified by estimating the relative pose between the candidate and the query KF.
The first step is handled by the Place Recognition module, which queries the KF Database for similar KFs based on the BoW \cite{dbow2} image descriptor.

The second step, the geometrical verification, is used as an additional check beside the visual appearance  and in order to obtain a transformation between the query KF $\text{KF}_q$ and the candidate KF $\text{KF}_c$.
Hence, with $q$ being the sensor coordinate frame of the $\text{KF}_q$ and $c$ being the coordinate frame of the sensor for $\text{KF}_c$, the goal is to find the transformation $T_{cq}$, describing the transformation from frame $q$ into frame $c$.
The standard way of computing this transformation is to establish 3D-2D correspondences between the query KF and the map points associated with the candidate KF. 
However, in COVINS-G the system does not have access to 3D map points, but only 2D keypoints.
Estimating the transformation between $\text{KF}_q$ and $\text{KF}_c$ using standard 2D-2D relative pose estimation algorithms like the 5-point algorithm \cite{5PT} would result in a scale ambiguity.
Instead, we propose to not only use the query and candidate KFs but also use some of their neighboring KFs and the relative odometry transformations to form two sets of cameras which can be treated as a multi-camera system as illustrated in Figure \ref{fig:17_PT}.
Using the 17-point algorithm \cite{17PT}, the transformation between two such camera systems can be estimated at metric scale.

\subsubsection{The 17-point Algorithm}
\label{sec:seventeen_pt}
\begin{figure}
    \centering
    \includegraphics[width=0.45\textwidth]{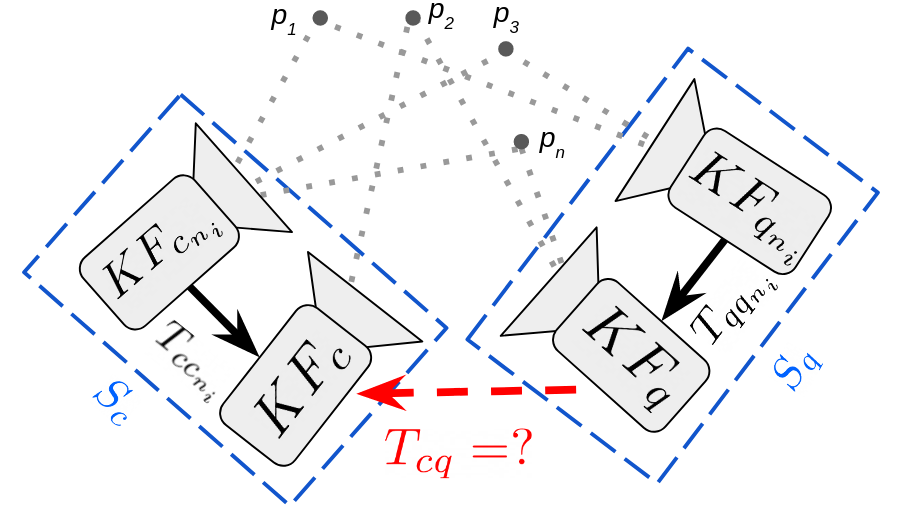}
    \caption{The 17-point algorithm requires 17 2D point correspondences to estimate the relative transformation $T_{cq}$ between the multi-camera systems $S_{c}$ and $S_{q}$ (represented by blue dashed lines). }
    \label{fig:17_PT}
    \vspace{-15pt}
\end{figure}

The 17-point (or 17-ray) algorithm \cite{17PT} can be used to solve for the relative motion between two multi-camera systems. 
A multi-camera system consists of more than one camera where the relative transformations between the cameras are known. 
Using 17 2D-2D point correspondences, the relative transformation between two viewpoints of a multi-camera system can be estimated. 
To do so, the generalized epipolar constraints \cite{generalized_epipolar} are used, which can be seen as an extension of the epipolar constraints to systems without a central point of projection.
Note that in our setup we only have a monocular camera, however, we leverage the fact that we have a good estimate of the relative pose between adjacent KFs provided by the VIO front-end.
Hence for a $\text{KF}_a$, we take the $m$ neighbors $\text{KF}_{a_{n_i}}, i \in [1, m]$ of $\text{KF}_a$ and use the relative transformations $T_{aa_{n_{i}}}$ to build up a conceptual multi-camera system (Figure \ref{fig:17_PT}).

\subsubsection{Implementation}
\label{sec:seventeen_pt_approach}
To estimate the transformation between the candidate KF ($\text{KF}_c$) and the query KF ($\text{KF}_q$), we extend the neighborhood to form two multi-camera systems $S_{q}$ and $S_{c}$ for the $\text{KF}_q$ and the $\text{KF}_c$, respectively.
In our setup the two sets comprised of one additional neighbor for $\text{KF}_q$ and two additional neighbors for $\text{KF}_c$, i.e. $S_{q} = \{ \text{KF}_q, \text{KF}_{q_{n_1}} \}$ and $S_{c} = \{ \text{KF}_c, \text{KF}_{c_{n_1}}, \text{KF}_{c_{n_2}}\}$.
In order to establish the 2D-2D candidate correspondences, we first perform a brute force descriptor matching across all pairs of the two sets $S_{q}$ and $S_{c}$, leading to 6 sets of correspondences.

Instead of using this set of correspondences directly with the 17-point algorithm inside a RANSAC framework, we first perform a pre-filtering, as a high outlier ratio would render a RANSAC with 17 correspondences infeasible.
The pre-filtering consists of performing a standard 2D-2D RANSAC on each of the 6 sets of correspondences.
To reject bad candidates early on, we require a minimum number of inliers (30) in each of the 6 sets.
On the remaining inlier correspondences, we carry out a 17-point RANSAC, where during the sampling we ensure that candidates from all sets are present in order to prevent degenerate configurations (e.g. all 17 correspondences from a single set).
For both the pre-filtering as well as the 17-point RANSAC, we utilize the OpenGV library \cite{opengv}.

For accepting a transformation we require to have at least 100 inliers after RANSAC.
To quantify the uncertainty of the computed transformation we use a sampling-based approach to compute a covariance matrix.
The samples are obtained by repeatedly selecting 17 inliers and computing the transformation using the 17-point algorithm.
We favor the sampling approach over an analytical one as it results in a more consistent covariance estimate since unmodelled effects (e.g. the relative pose uncertainty of the odometry) are better reflected in the samples.

\subsection{Pose Graph Optimization}
\label{sec:pgo}
The PGO in COVINS-G is where the information from the different agents gets fused and the drift in the corresponding trajectories can be corrected.
The PGO step is triggered every time a new constraint is added to the graph as a result of the loop detection.
The state that is optimized in the PGO consists of all KF poses that are in the corresponding graph. 
In the following, we denote the pose of KF $k$ by a rotation $q_{ws_k}$ and a translation $p_{ws_k}$, where $w$ represents the coordinate frame in which the poses are expressed and $s_k$ denotes the device coordinate frame at time-point $k$.
Hence, the state in PGO can be defined as $\mathcal{S} = \{ q_{ws_1}, p_{ws_1}, \cdots, q_{ws_n}, p_{ws_n} \}$, with $n$ being the number of keyframes in the graph.

In the actual PGO, we optimize the following objective:
\begin{equation}
\label{eq:pg}
   \mathcal{S}^{*} = \underset{\mathcal{S}}{\text{arg min}} \{ \sum_{k=1}^{n}  \sum_{l=1}^{q} \norm{e_{kk+l}}^2_{W_{kk+l}} + \sum_{i,j \in \mathcal{L}} \varrho (\norm{e_{ij}}^2_{W_{ij}}) \},
\end{equation}
where $\norm{e}^2_{W} = e^TWe$ is the squared Mahalanobis distance with the information matrix $W$, $\mathcal{L}$ denotes the set of all loop-closure edges and $\varrho(\cdot)$ denotes the use of a robust loss function, here, the Cauchy loss function.
The error terms $e_{ij}$ represent the standard relative pose residual between KF $i$ and $j$.
The parameter $q$ represents the number of neighbor KFs added as odometry constraints (e.g. $q=1$ would correspond to having an edge only with the subsequent KF). 
This is used to approximate correlations between poses which are generally computed in a sliding window fashion and in our implementation is set $q=4$.
%
%
%
%
For the optimization, the Ceres's implementation of the Levenberg-Marquardt is used.
The information matrices $W$ for the loop constraints are obtained via the estimated covariance outlined in section \ref{sec:seventeen_pt_approach}.
The ones corresponding to the odometry edges are obtained using the expected accuracy of the corresponding front-end.


\section{Experiments and Discussions}
\label{sec:experiments}
In our evaluation, we demonstrate the generic nature of our collaborative back-end and its capability to use and combine all sorts of VIO front-ends.
%
%
All our experiments are performed using pre-recorded datasets which we play back in real-time.
The agents' front-ends are run on Intel NUC 7i7BNH @3.5 GHz hardware, while the server is run on a laptop with a core i7-6700HQ @2.6 GHz.
Note that in our setup the agents and the server are connected via a wireless network, in order for real communication to take place.
This setup allows us to have more comparable results over multiple runs while still making use of a real wireless network as would be the case during real-world deployment.

\begin{table}[!htb]
\caption{Comparison of all the loops found using five agents on the EuRoC MH datasets between the map point based PnP and the 17Pt algorithm. The accuracies are reported with their corresponding standard deviations.}
\label{tab:loop_trans}
\begin{adjustbox}{max width=\columnwidth}
\begin{tabular}{|l|l|l|l|l|}
\hline
          \bf{Front-end}                 & \bf{Method} & \bf{ \#Loops} & \bf{Transl. Err. (m)} & \bf{Rot. Err. (deg)} \\ \hline
\multirow{2}{*}{VINS-mono} & PnP    & 852     & $0.69 \pm 0.6$   & $3.44 \pm 3.5$              \\
                           & 17Pt   & \bf{1823}    & $\mathbf{0.10 \pm 0.17}$ &  $\mathbf{1.09 \pm 1.25}$           \\ \hline
\multirow{2}{*}{ORB-SLAM3} & PnP    & \bf{505}     & $0.23 \pm 0.92$  & $1.56 \pm 3.7$     \\
                           & 17Pt   & 406     & $\mathbf{0.07 \pm 0.12}$ & $\mathbf{1.01 \pm 0.9}$          \\ \hline
\end{tabular}
\end{adjustbox}
\vspace{-8pt}
\end{table}

\subsection{Loop Transformation Evaluation}
\label{sec:loop_accuracy}
To evaluate the quality of the loop transformations, we compare the estimated loop transformations using the 17Pt algorithm against the PnP based method used in COVINS on the EuRoC Machine Hall (MH) dataset\cite{euroc}.
As shown in Table \ref{tab:loop_trans}, for both the VINS-mono as well as the ORB-SLAM3 front-end, using the 17Pt algorithm achieves a higher accuracy compared to the map point based PnP method. 
Furthermore, it can be seen that the accuracy of our proposed approach does not suffer from noisy map points like is the case with VINS-mono, and also is able to find significantly more successful loops in such cases.

As not only the accuracy of the loops is important, but also an accurate measure of the uncertainty to be used as a weight in PGO, we evaluated the Normalized Estimation Error Squared (NEES) \cite{nees} for all successful loops recorded in Table \ref{tab:loop_trans}.
Using our proposed sampling-based covariance estimation scheme, we achieve an average NEES of roughly 30.
Estimating the covariance using the residual formulation in the 17Pt algorithm in an analytical fashion results in a NEES which is several magnitudes larger.
While our approach is still slightly overconfident, the estimated covariance is significantly more accurate compared to the analytical approach, which does not take into account effects such as drift and noise on the estimated poses.

\subsection{Collaborative SLAM Estimation Accuracy}
\label{sec:euroc_eval}

We evaluate the accuracy of the collaborative SLAM estimate on the EuRoC Dataset\cite{euroc} using the Machine Hall (MH) and the Vicon Room 1 (V1) sequences to establish a collaborative estimation scenario with three to five participating agents. 
We use various combinations of front-ends with our back-end and compare the results against COVINS \cite{covins} as well as the state-of-the-art multi-session capable VINS-mono\cite{vins-mono} and  ORB-SLAM3 \cite{ORBSLAM3_TRO}.
As multi-session methods only support one agent at a time, we process the datasets sequentially with one agent at a time, in contrast to the collaborative SLAM approaches, where all agents' processes are run in parallel.

To showcase the flexibility of our approach, we perform the evaluation of our back-end with different front-end combinations.
In the first experimental setup, all agents are operated using the same front-end, which in our experiments is either the VINS-mono or the ORB-SLAM3 front-end.
In subsequent experiments, we test COVINS-G using different front-ends for the different agents, namely OpenVINS\cite{Openvins}, VINS-mono, ORB-SLAM3 and SVO-pro\cite{Forster17troSVO}.
Since COVINS performs as a post-processing step a Global Bundle Adjustment (GBA) at the end of every run, we do not directly compare against it but rather include its performance here as a reference.
For more appropriate comparisons with COVINS-G, we include the COVINS result without the additional GBA step.
The averaged results over 5 runs are summarized in Table \ref{tab:multi_ag}.

Using the ORB-SLAM3 front-end, we achieve on-par performance to the ORB-SLAM3 multi-session back-end, even though in our approach we do not perform map re-use as in ORB-SLAM3.
As ORB-SLAM3 performs a GBA upon every loop detected, it can potentially reach very high accuracy (as COVINS demonstrates with GBA enabled), however, as long as it is able to find sufficient correspondences in the re-localization, no GBA gets triggered.
Therefore, the overall accuracy is coupled also to the time that the last GBA step was performed.
Compared to COVINS without GBA, our approach reduces the error almost by a factor of two.
This can be explained by the fact that in COVINS-G we are able to close more loops, as also shown in Table \ref{tab:computation}.

The comparison with the VINS-mono front-end showcases a similar outcome that COVINS-G performs similarly to the VINS-mono multi-session back-end.
The small gap in performance can be explained in that VINS-mono incorporates a larger number of loops, whereas in COVINS-G we set a minimum number of KFs between consecutive loops in order to reduce the computational burden.
Compared to COVINS with the VINS-mono frontend, COVINS-G shows a significant improvement with a factor of around 3.
This is because COVINS only detects and closes few loops because it's loop-closure pipeline is tailored to a high quality map with a large number of map points.
Due to this weak connection within and across the trajectories, even the GBA is unable to reduce the error.
An improvement in performance is also observed by utilizing the covariance of the loop estimates for weighting the edges during PGO as opposed to using a fixed weighting scheme as done in VINS multi-session.
With COVINS-G, even with a mix of different front-ends, we can see that the performance is in a similar range as when using the VINS-mono front-end, demonstrating the effectiveness of our generic approach.

\begin{table}[htb!]
\caption{Evaluation of joint-trajectory estimates for different methods on the EuRoC Dataset\cite{euroc} (lowest error in bold) reported as the average trajectory error over 5 runs each. $^\#$For the heterogenous front-end, agents 1-5 utilize OpenVINS\cite{Openvins}, VINS-mono\cite{vins-mono}, ORB-SLAM3\cite{ORBSLAM3_TRO}, SVO-Pro\cite{Forster17troSVO} and VINS-mono\cite{vins-mono} front-ends, respectively. As *COVINS\cite{covins} performs a Global Bundle Adjustment (GBA) step at the end of the run it has been included here for reference only and is excluded from our comparison.
}
\label{tab:multi_ag}
\centering
\def\arraystretch{1.2}
\begin{adjustbox}{max width=\columnwidth}
\begin{tabular}{|cc|ccc|}
\hline
\multicolumn{2}{|c|}{\textbf{Method}}                                                          & \multicolumn{3}{c|}{\textbf{Translational RMSE (m)}}                                                                                 \\ \hline
\multicolumn{1}{|c|}{}                                 &                                       & \multicolumn{1}{c|}{MH01-MH03}                    & \multicolumn{1}{c|}{MH01-MH05}                    & V101-V103                    \\ \cline{3-5} 
\multicolumn{1}{|c|}{\multirow{-2}{*}{Front-end}}       & \multirow{-2}{*}{Back-end}             & \multicolumn{1}{c|}{(3 Agents)}                   & \multicolumn{1}{c|}{(5 Agents)}                   & (3 Agents)                   \\ \hline

\multicolumn{1}{|c|}{ORB-SLAM3\cite{ORBSLAM3_TRO}}                        & ORB-SLAM3 MS\cite{ORBSLAM3_TRO}                          & \multicolumn{1}{c|}{0.041}                        & \multicolumn{1}{c|}{0.082}                        & \textbf{0.048}               \\ \hline
\multicolumn{1}{|c|}{ORB-SLAM3\cite{ORBSLAM3_TRO}}                        & COVINS (No GBA) \cite{covins}                      & \multicolumn{1}{c|}{0.075}                        & \multicolumn{1}{c|}{0.119}                        & 0.130                        \\ \hline
\multicolumn{1}{|c|}{{\color[HTML]{666666} ORB-SLAM3\cite{ORBSLAM3_TRO}}} & {\color[HTML]{666666} COVINS (GBA) \cite{covins}*} & \multicolumn{1}{c|}{{\color[HTML]{666666} 0.024}} & \multicolumn{1}{c|}{{\color[HTML]{666666} 0.036}} & {\color[HTML]{666666} 0.042} \\ \hline
\multicolumn{1}{|c|}{ORB-SLAM3\cite{ORBSLAM3_TRO}}                        & Ours                                  & \multicolumn{1}{c|}{\textbf{0.040}}               & \multicolumn{1}{c|}{\textbf{0.064}}               & 0.067                        \\ \hline \hline

\multicolumn{1}{|c|}{VINS-mono \cite{vins-mono}}                        & VINS MS\cite{vins-mono}                               & \multicolumn{1}{c|}{\textbf{0.062}}                        & \multicolumn{1}{c|}{0.100}                        & \textbf{0.076}                        \\ \hline
\multicolumn{1}{|c|}{VINS-mono \cite{vins-mono}}                        & COVINS (No GBA) \cite{covins}                               & \multicolumn{1}{c|}{0.259}                        & \multicolumn{1}{c|}{0.305}                        & 0.183                       \\ \hline
\multicolumn{1}{|c|}{{\color[HTML]{666666} VINS-mono \cite{vins-mono}}}                        & { \color[HTML]{666666} COVINS (GBA) \cite{covins}*}                               & \multicolumn{1}{c|}{ {\color[HTML]{666666} 0.261}}                        & \multicolumn{1}{c|}{{\color[HTML]{666666} 0.321}}                        & {\color[HTML]{666666} 0.183}                       \\ \hline
\multicolumn{1}{|c|}{VINS-mono\cite{vins-mono}}                        & Ours w/o Covariance                              & \multicolumn{1}{c|}{0.092}                        & \multicolumn{1}{c|}{0.110}                        & 0.117                        \\ \hline
\multicolumn{1}{|c|}{VINS-mono\cite{vins-mono}}                        & Ours                                  & \multicolumn{1}{c|}{0.081}                        & \multicolumn{1}{c|}{\textbf{0.095}}                        & 0.090                        \\ \hline \hline

\multicolumn{1}{|c|}{Heterogenous$^\#$}                   & Ours                                  & \multicolumn{1}{c|}{0.081}                        & \multicolumn{1}{c|}{0.090}                        & 0.088                        \\ \hline
\end{tabular}
\end{adjustbox}
\vspace{-10pt}
\end{table}

\subsection{Large-Scale Outdoor Experiments}
\label{sec:outdoor_expt}
\begin{figure}[h]
    \centering
    \includegraphics[width=0.48\textwidth]{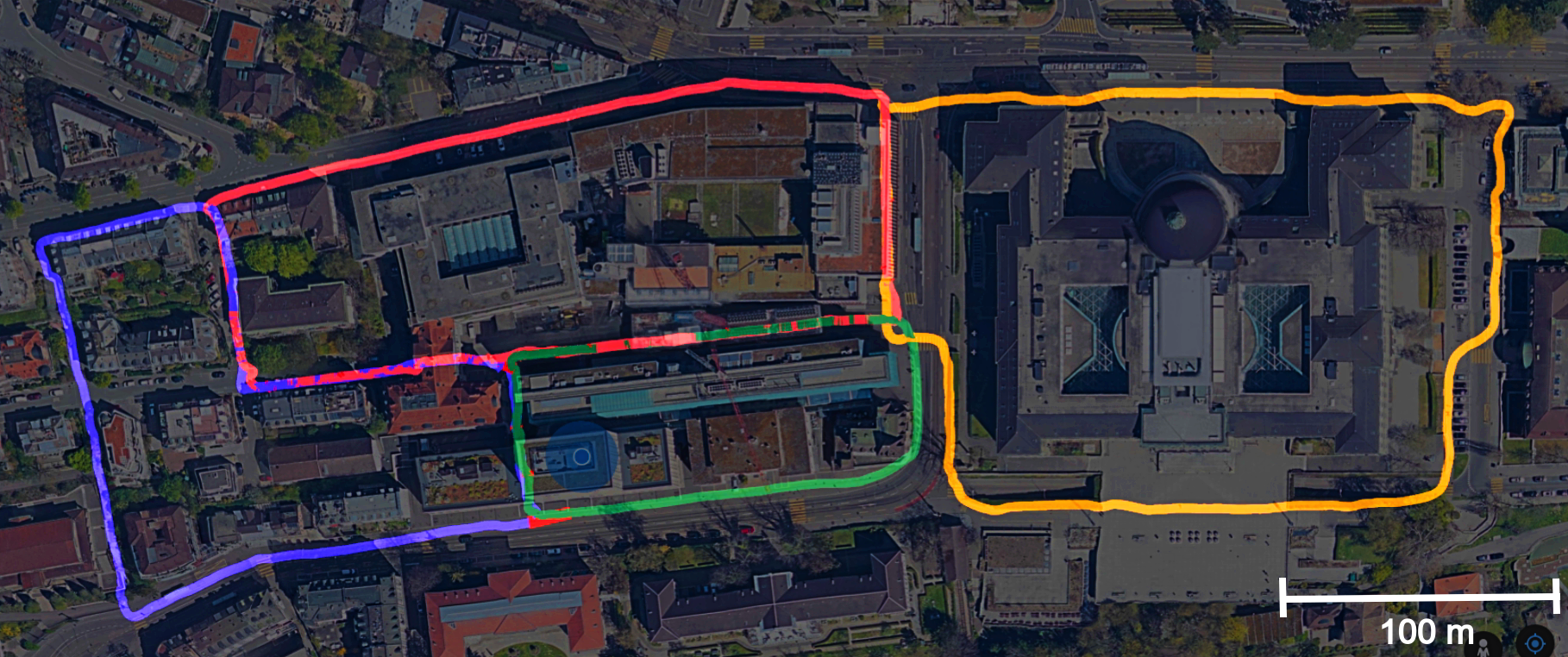}
    \caption{Joint trajectory estimates for 4 agents superimposed on the satellite image of our large-scale outdoor experiment with a combined trajectory length of 2400 $m$.}
    \label{fig:outdoor_expt}    
    \vspace{-12pt}
\end{figure}

To demonstrate the applicability of our back-end to large-scale scenarios, we captured a dataset consisting of 4 sequences using a hand-held setup with an Intel Realsense D455.
The dataset was recorded by walking on the sidewalks around the campus of ETH Zurich in the center of Zurich and with a combined trajectory length of around 2400$m$ covering an area of approximately 67500$m^2$.
For this experiment, the VINS-mono front-end was used on all sequences.
As we do not have a ground truth trajectory for the dataset, we superimposed the estimated combined trajectory on a satellite image of the area as illustrated in Figure \ref{fig:outdoor_expt}. 
As can be seen, the different trajectories are well aligned with the road structure. 
A complete visualization of the whole experiment can be found in the supplementary video.

\subsection{Tracking Camera Support}
\label{sec:tracking_cam}
This experiment aims at highlighting the generalization capabilities of our system with respect to the front-end that is used.
To do so, we captured a dataset with two different sensors, one Intel Realsense D455 camera, and one Intel Realsense T265 tracking camera inside an office floor.
For the D455 we used the VINS-mono as a front-end, whereas for the T265 we used the odometry estimate that is provided directly by the sensor.
As this sensor only provides its images and the corresponding poses but does not offer access to its internal state, we use our ROS-based front-end wrapper, which does a motion-based keyframe selection, detects feature points, and creates and communicates the KF messages to the back-end.

\begin{figure}[h]
    \centering
    \includegraphics[width=0.48\textwidth]{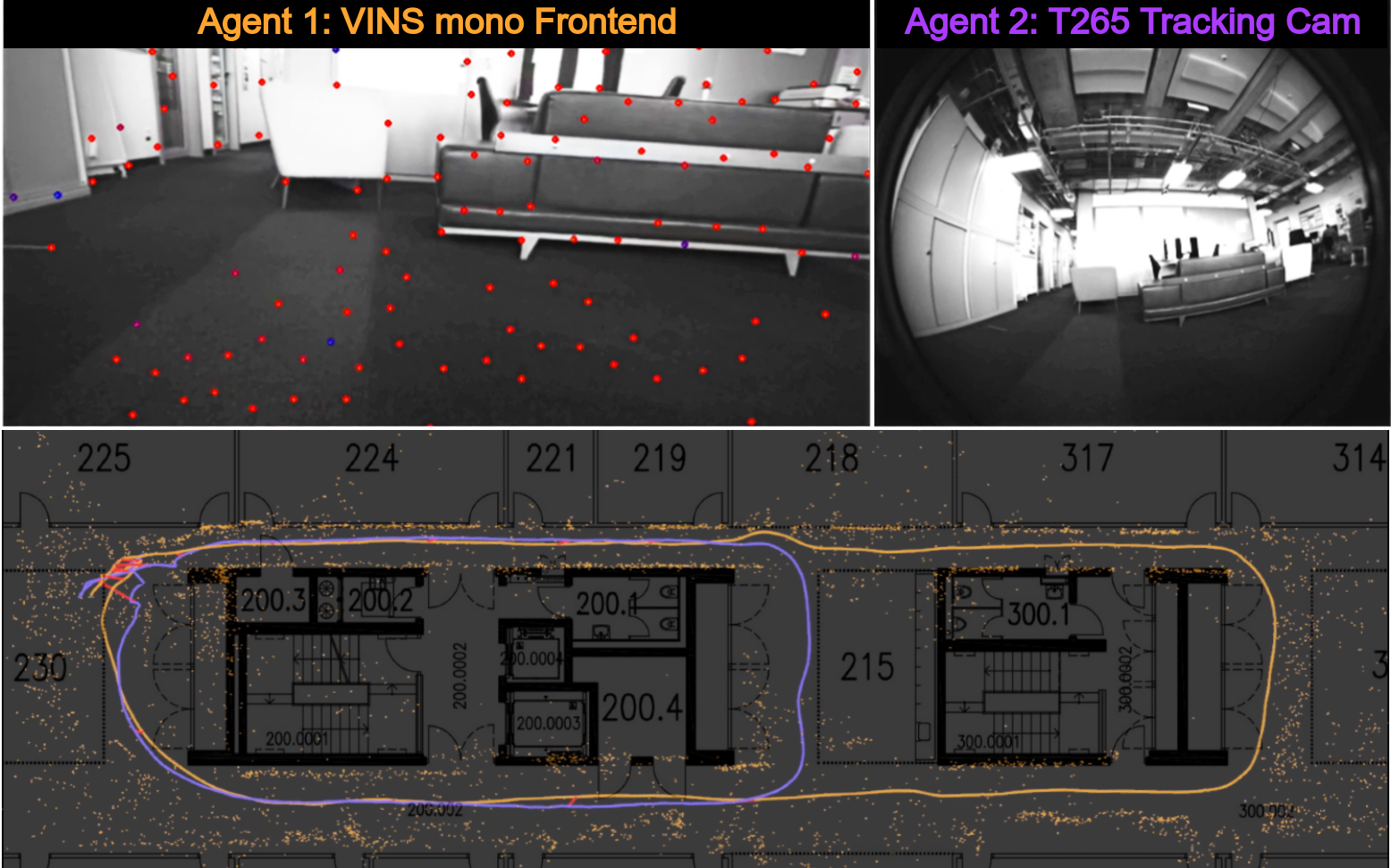}
    \caption{Joint trajectory estimate for a heterogeneous system of 2 agents superimposed on the floor plan of the 
    building. Map points are shown for visualization purposes only}
    \label{fig:tracking}    
    \vspace{-9pt}
\end{figure}

The outcome of the experiment is illustrated in Figure \ref{fig:tracking}, where one can see an example image for each of the sensors and the combined trajectory overlaid with the floor plan of the office (the complete experiment is visualized in the supplementary video).
As it can be seen, the estimated trajectory fits the floor plan, indicating its accuracy.
In addition to the different odometry sources, the two sensors have also two rather different lenses, rendering matching between the two systems more difficult, however, our proposed back-end is able to merge the estimates nonetheless.

\subsection{Alternative Feature Descriptor support}
\label{sec:feature_support}
Visual features are designed to be tolerant to illumination changes, perspective distortions as well as viewpoint changes.
While for runtime efficiency, real-time SLAM systems are mainly restricted to highly computational efficient binary features such as ORB \cite{ORB} or BRISK \cite{Leutenegger:etal:ICCV2011}.
However, it is accepted that to date, binary features are not able to match the robustness of more expensive descriptors such as e.g. SIFT \cite{SIFT}.
As our approach is largely decoupled from the front-end, we are able to make use of more expensive features, also because we do not require them to be computed at frame rate, thus allowing us to potentially match trajectories with much larger viewpoint differences.

In this experiment, we recorded another outdoor dataset with two hand-held trajectories walking in parallel on two sides of a street.
We modified our front-end wrapper to detect SIFT instead of ORB features and run the back-end using these SIFT features as well.
While using the framework with ORB features no overlap can be detected to merge the trajectories, using the powerful SIFT features allowed us to successfully detect loops across the agents and align the trajectories.
The comparison of this experiment can be found in the accompanying video.
Such an improved tolerance to large view-point changes allows the system to be used in scenarios where larger view-point changes are inherent to the use case, for example, the collaboration between aerial and ground robots.

\subsection{Computational and Communication Requirements}
\label{sec:computation}

The statistics for communication and loop transformation computation are generated and compared with COVINS for the experiment performed on EuRoC MH dataset with 5 agents, each running an ORB-SLAM3 front-end. 
The computation time for estimating the loop transformation using our approach (including feature matching, 2D-2D pre-filtering, 17-point RANSAC and covariance matrix computation) is compared against the standard approach used in COVINS (feature matching + PnP RANSAC) and summarized in Table \ref{tab:computation}. 
Though the computation time for our method is around an order of magnitude higher than for COVINS, the average computation time per loop is around $213 ms$ which still makes it possible to operate in real-time for up to 5 loop computations per second. 
The network traffic for our approach is around three times lower than that of COVINS.
This is owed to the fact that in our back-end we require only 2D keypoints and their descriptors for each KF, unlike COVINS which also requires sending the map points as well as the data required to perform the IMU pre-integration.

\begin{table}[htb!]
\caption{Comparison of the runtime for the loop computation and the average communication traffic.}
\label{tab:computation}
\centering
\def\arraystretch{1.2}
\begin{tabular}{|c|c|c|c|}
\hline
\multicolumn{1}{|l|}{\textbf{}} & \textbf{\begin{tabular}[c]{@{}c@{}}\# Loops\\  detected\end{tabular}} & \textbf{\begin{tabular}[c]{@{}c@{}}Computation Time \\ per loop\end{tabular}} & \textbf{\begin{tabular}[c]{@{}c@{}}Network Traffic\\  per Agent\end{tabular}} \\ \hline
Ours                            & 75                                                                    & 213 $\pm$ 171 ms                                                                        & 179.74 kB/s                                                                   \\ \hline
COVINS \cite{covins}                         & 57                                                                    & 35 $\pm$ 23 ms                                                                         & 486.41 kB/s                                                                   \\ \hline
\end{tabular}
\vspace{-15pt}
\end{table}


\section{Conclusion}
\label{sec:conclusion}
In this work, we present a front-end-agnostic back-end for collaborative visual-inertial SLAM.
By making use of a multi-camera relative pose estimator for estimating loop-closure transformations, our system is able to work using only 2D keypoint information.
This allows our collaborative back-end to be compatible with virtually any VIO front-end with minimal to no modifications to it.
In our experimental evaluation, we achieve at least on-par accuracy with state-of-the-art multi-session and collaborative SLAM systems, which use back-ends specifically designed for the respective front-end.
Owed to the decoupled nature of the data used in our back-end, we can make use of more powerful keypoint descriptors like SIFT, allowing us to close loops and merge trajectories with drastic viewpoint changes, which cannot be handled by state-of-the-art systems.
Our open-source implementation of the system enables users to unlock the capabilities of a collaborative SLAM system without the need to change their state estimation pipeline of choice.
In the future we would like to exploit the fact that the VIO estimates are gravity aligned and replace the 17-point algorithm with the minimal 4Pt solver \cite{4PT_gravity}, allowing to speed up the RANSAC computation as fewer samples are required.

%
%







\clearpage
\bibliographystyle{IEEEtran}
\bibliography{ref}

\end{document}